\documentclass{article}

\PassOptionsToPackage{numbers}{natbib}
   \usepackage[final]{neurips_2019}

\usepackage[utf8]{inputenc} 
\usepackage[T1]{fontenc}    
\usepackage{hyperref}       
\usepackage{url}            
\usepackage{booktabs}       
\usepackage{amsthm}
\usepackage{graphicx}
\usepackage{textcomp}
\usepackage{amsmath}
\usepackage{amssymb}
\usepackage{microtype}
\usepackage{subcaption}
\newcommand{\rpm}{\raisebox{.2ex}{$\scriptstyle\pm$}}
\usepackage{amsfonts}       
\usepackage{nicefrac}       
\usepackage{microtype}      
\usepackage{fancyhdr}
\theoremstyle{definition}
\newtheorem{definition}{Definition}
\usepackage{bbm}
\usepackage[font=small,labelfont=bf]{caption}
\usepackage{tabulary, booktabs, subcaption}
\title{Accurate Layerwise Interpretable Competence Estimation}

\author{Vickram Rajendran, William LeVine \\The Johns Hopkins University Applied Physics Laboratory\\Laurel, MD 20723 \\ \texttt{\{vickram.rajendran, william.levine\}@jhuapl.edu}}
\begin{document}

\maketitle
\begin{abstract}

Estimating machine learning performance “in the wild” is both an
important and unsolved problem. In this paper, we seek to examine,
understand, and predict the pointwise competence of classification models. Our
contributions are twofold: First, we establish a statistically
rigorous definition of competence that generalizes the common notion of classifier confidence; second, we present the ALICE (Accurate Layerwise Interpretable Competence Estimation)
Score, a pointwise competence estimator for any classifier.  By considering
\emph{distributional, data,} and \emph{model uncertainty}, ALICE empirically
shows accurate competence estimation in common failure situations
such as class-imbalanced datasets, out-of-distribution datasets, and
poorly trained models.

Our contributions allow us to accurately predict the competence of any
classification model given any input and error function. We compare
our score with state-of-the-art confidence estimators such as
model confidence and Trust Score, and show significant improvements in
competence prediction over these methods on datasets such as DIGITS, CIFAR10, and CIFAR100. 
\end{abstract}

\section{Introduction}
\label{sec:org9894a65}

Machine learning algorithms have achieved tremendous success in areas
such as classification \citep{alexnet}, object detection \citep{yolov3}, and
segmentation \citep{deeplab}. However,
as these algorithms become more prevalent in society it is essential
to understand their limitations. In particular, a supervised machine
learning model's performance on a reserved test point is characterized by the difference between that point's label and the model's prediction on that point. A model is considered performant on that point if this difference is sufficiently small; unfortunately, this difference is impossible to compute
once the model is deployed since the point's true label is unknown. 

This problem is exacerbated when we consider the difference between
real world data and the curated datasets that the
models are evaluated on --- often these datasets are significantly
different, and it is not clear whether performance on a held aside
test set is indicative of real-world performance. It is essential to
have a predictive measure of performance that does not require ground
truth in order to determine whether or not a machine learning
algorithm's prediction should be trusted "in the wild" --- a measure of model
competence. However, competence is currently not defined in any
rigorous manner and is often restricted to the more specific idea of model
confidence.

In this paper, we define competence to be a generalized form of predictive uncertainty, and so we must account for all of its' generating facets. Predictive uncertainty arises from three factors: \emph{distributional,
data}, and \emph{model uncertainty}. \emph{Distributional uncertainty}
\citep{Gal2016Uncertainty} arises from mismatched training and test
distributions (i.e. dataset shift \citep{dataset_shift}). \emph{Data
uncertainty} \citep{Gal2016Uncertainty} is inherent in the complex
nature of the data (e.g. input noise, class overlap, etc.). Finally,
\emph{model uncertainty} measures error in the approximation of the true
model used to generate the data (e.g. overfitting, underfitting,
etc.)  \citep{Gal2016Uncertainty} --- this generally reduces as the amount of
data increases. Accurate predictive uncertainty estimation (and thus accurate competence estimation) requires consideration of all three
of these factors. Previous attempts to explicitly model these three factors require out-of-distribution data, or are not scalable to high dimensional datasets or deep networks \citep{Gal2016Uncertainty} \citep{NIPS2018_7936}; there are currently very few methods that
do so in a way that requires no additional data, scales to high
dimensional data and large models, and applies to any classification model,
regardless of architecture, dataset, or performance.

We focus on mitigating these issues in the space of
classifiers. In Section \(2\) we present several definitions, including
a robust, generalizable definition of model \emph{competence} that encompasses the common notion of model confidence. In Section \(3\) we examine the related work in the areas
of predictive uncertainty estimation and interpretable machine
learning. In Section \(4\) we show a general metric for evaluating
competence estimators. In Section \(5\) we develop the "ALICE Score," an accurate
layerwise interpretable competence estimator. In Section \(6\) we
empirically evaluate the ALICE Score in situations involving different types of predictive
uncertainty and on various models and datasets. We
conclude in Section \(7\) with implications and ideas for future work.
\section{Definitions}
\label{sec:orgf6ca9a3}
\begin{definition}
  \label{error} \textbf{(Error Function)} Let \(\mathcal{C}\) be the finite "label space" of
possible labels of the true model \(f\) used to generate data, and let $\mathcal{Y}$ be the associated unit simplex of class probabilities, which we call the "distributional space". Let
\(\hat{\mathcal{Y}} \subseteq \mathcal{Y}\) be the space of possible outputs of a classifier
\(\hat{f}\) that approximates \(f\). We will denote the classes in $\mathcal{C}$ predicted by these models (usually through an argmax) as $\hat{f}_c$ and $f_c$. An error function \(\mathcal{E}\) is a
function \(\mathcal{E}\colon \mathcal{Y} \times \hat{\mathcal{Y}} \to
\mathbb{R}^{\geq 0} \cup \{+\infty\}\), with the property that
\(\mathcal{E}(y, \hat{y}) = \infty\) when \(y \in 
\overline{\mathcal{\hat{Y}}} \cap \mathcal{Y}\). This property intuitively means that the output of the error function is infinite if the true class is outside of the classifier's prediction space. Given a point \(x\), we denote
\(\mathcal{E}(f(x), \hat{f}(x))\) the error of \(\hat{f}\) on \(x\). Common
examples of error functions are mean squared error, cross-entropy error, and
0-1 error (the indicator that the classes predicted by \(\hat{f}\) and
\(f\) are different). Note that an error function is distinct from a
loss function since it is neither required to be differentiable nor
continuous.
\end{definition}
\begin{definition}
\label{confidence} \textbf{(Confidence)} The commonly accepted definition of
classifier confidence \citep{mandelbaum2017distance}
\citep{chen2018confidence} \citep{NIPS2018_7936} \citep{uncertainty_conf}
is the probability that the model's predicted class on an input \(x\) is
the true class of \(x\). Explicitly, this is \(p(f_c(x) = \hat{f}_c(x) | x,
\hat{f})\). This is also the inverse of the \textbf{predictive
uncertainty} \citep{uncertainty_conf} of a classifier, which is the
probability that the model's prediction is incorrect
\citep{NIPS2018_7936}.
\end{definition}

  While confidence is sufficient in many cases, we would
like to have a more general and flexible definition that can be tuned
towards a specific user's goals. For example, some users may be
interested in top-\(k\) error, cross-entropy or mean
squared error instead of 0-1 error. We can model this by
rewriting the confidence definition with respect to an error function
\(\mathcal{E}\):
\begin{align*}
p(f_c(x) = \hat{f}_c(x) | x, \hat{f}) &= p(\mathcal{E}(f(x), \hat{f}(x)) = 0 | x, \hat{f})
\end{align*}
where \(\mathcal{E}\) is the 0-1 error. We can now extend
 \(\mathcal{E}\) beyond $\mathcal{E}_{0-1}$ to fit an end-user's goals. We can make this definition
 even more general by borrowing ideas from the Probable Approximately Correct
 (PAC) Learning framework \citep{pac_valiant} and allowing users to specify
 an error tolerance \(\delta\). For example, some users may allow for
 their prediction error to be below a specific \(\delta\) for their model to be considered competent. One could imagine that for highly precise problems with low threshold for error, $\delta$ would be quite low, while less stringent use-cases could allow for larger $\delta$'s. The relaxation of the prediction error leads to the generalized notion of \textbf{$\delta$-competence}, which we define as $p(\mathcal{E}(f(x), \hat{f}(x)) < \delta | x, \hat{f})$. Confidence can be recovered by setting $\mathcal{E} = \mathcal{E}_{\text{0-1}}$ and $\delta \in (0, 1)$. 

Allowing both \(\delta\) and \(\mathcal{E}\) to vary
 gives fine control to an end-user about the details of a model's
 performance with respect to a specific error function.
\begin{definition}
\label{competence} \textbf{(\(\delta\)-\(\epsilon\) Competence)} The
   \emph{true} \textbf{\(\delta\)-competence} of a model  at a
   given point is the binary variable \(\mathcal{E}(f(x), \hat{f}(x)) < \delta | x, f, \hat{f})\) where $\mathcal{E}$ is an error function (Definition \ref{error}). Note that \(\mathcal{E}\) becomes a random variable when \(f\) is unknown since $\mathcal{E}$ is a deterministic function of
   the uncertain variable \(f(x)\) --- this notion of randomness is slightly distinct from treating $\hat{f}$ as a random variable due to finite data. Given that \(f\) is unknown, we must estimate the \textbf{\(\delta\)-competence}, which can now be written as $p(\mathcal{E}(f(x), \hat{f}(x)) < \delta | x, \hat{f})$. Putting a risk threshold \(\epsilon\) on the value of the \(\delta\)-competence leads us to the following notion: A model is
   \textbf{\(\delta\)-\(\epsilon\) competent} with respect to \(\mathcal{E}\)
   at \(x\) if \(p(\mathcal{E}(f(x), \hat{f}(x)) < \delta | x, \hat{f}) >
   \epsilon\), or it is likely to be approximately correct. 
\end{definition}

This definition of competence allows a user to set a correctness
threshold (\(\delta\)) on how close the prediction and the true output
need to be in order to be considered approximately correct, as well as
set a risk threshold (\(\epsilon\)) on the probability that this
prediction is approximately correct with respect to any error function. These thresholds and error
functions allow for a flexible definition of competence that can be
adjusted depending on the application. This also follows the
definition of trust in \citep{Lee_trust} as "the attitude that an agent
will help achieve an individual’s goals in a situation characterized
by uncertainty and vulnerability."

Since we neither have access to labels nor have enough information to efficiently compute the true probability distribution $p(\mathcal{E}(f(x), \hat{f}(x)) < \delta | x, \hat{f})$ we seek to estimate this probability. We make this clear with the following definition:

\begin{definition} 
\label{comp_est} \textbf{(Competence Estimator)} A competence estimator of a
  model \(\hat{f}\) with respect to the error function \(\mathcal{E}\) is
  a function \(g_{\hat{f}}\colon \mathcal{X} \times \mathbb{R} \to [0, 1]\),
  where \(\mathcal{X}\) is the space of inputs, that is a statistical
  point estimator of the true variable \(\mathcal{E}(f(x),
  \hat{f}(x)) < \delta | x, \hat{f}, f\). In particular,
  \(g_{\hat{f}}(x, \delta) = \hat{p}(\mathcal{E}(f(x), \hat{f}(x)) < \delta | x, \hat{f})\).
\end{definition}

In the future we omit conditioning on \(\hat{f}\) in our
notation with the note that all subsequent probabilities are
conditioned on \(\hat{f}\).
\section{Related Work}
\label{sec:org8363798}

Competence estimation is closely tied with the well-studied areas of
  predictive uncertainty and confidence estimation, which can
  further be divided into Bayesian approaches such as
  \citep{bayes_kendall} \citep{bayes_hinton} \citep{bayes_net_1}, or
  non-Bayesian approaches including \citep{gal2015dropout},
  \citep{Oberdiek2018ClassificationUO},
  \citep{lakshminarayanan2017simple}. Bayesian methods attempt to
  determine some distribution about each of the weights in a network
  and predict a distribution of outputs using this distribution of
  weights. Computing the uncertainty of a prediction then becomes
  computing statistics about the estimated output
  distribution. These estimates tend to perform well, but tend not
  to be scalable to high dimensional datasets or larger
  networks. The non-bayesian methods traditionally fall under
  ensemble approaches \citep{lakshminarayanan2017simple}, training on
  out-of-distribution data \citep{NIPS2018_7936}
  \citep{Oberdiek2018ClassificationUO}
  \citep{Subramanya2017ConfidenceEI}, or dropout
  \citep{gal2015dropout}. This field tends to only work on a certain
  subset of classifiers (such as models with dropout for
  \citep{gal2015dropout}) or require modifications to the models in
  order to compute uncertainty \citep{mandelbaum2017distance}. Many
  of these methods are based off of the unmodified model confidence
  \citep{gal2015dropout}, and thus could be supplementary to our new
  competence score. To the best of our knowledge there are no
  existing Bayesian or non-Bayesian methods that consider competence with respect to error
  functions other than 0-1 error nor methods that have
  tunable tolerance parameters.

Another related area of research is interpretable machine
learning. Methods such as prototype networks
\citep{snell_prototype} or LIME \citep{ribeiro_lime} are very useful
in explaining why a classifier is making a prediction, and we
expect these methods to augment our work. However, competence
prediction does not attempt to explain the predictions of a
classifier in any way---we simply seek to determine whether or not
the classifier is competent on a point, without worrying about why or how the model made that decision. In this sense we are
more closely aligned with \emph{calibration} \citep{calibration_guo},
which adjusts prediction scores to match class conditional
probabilities which are interpretable scores
\citep{Subramanya2017ConfidenceEI} \citep{isotonic} and works such as \citep{rudin2019stop} are orthogonal to ours. While our goal is not
to compute class probabilites, our method similarly provides an
interpretable probability score that the model is competent.

The closest estimators to our own are \citep{chen2018confidence}
and \citep{jiang2018trust}. \citep{chen2018confidence} learns a meta
model that ensembles transfer classifiers' predictions to predict
whether or not the overall network has a correct
classification. Conversely, \citep{jiang2018trust} computes the
ratio of the distance to the predicted class and the second
highest predicted class as a Trust Score. While
\citep{chen2018confidence} takes into account \emph{data
uncertainty} with transfer classifiers, it does not explicitly take into account \emph{distributional} or \emph{model uncertainty}. Oppositely, \citep{jiang2018trust} considers neither
\emph{model} nor \emph{data uncertainty} explicitly, though it does model
\emph{distributional uncertainty} similarly to
\citep{lakshminarayanan2017simple}, \citep{lee2017training}, and
\citep{lee2018simple}. Further, both merely \emph{rank} examples according to
uncertainty measures that are not human-interpretable. They also
focus on \emph{confidence} rather than \emph{competence}, which does not
allow them to generalize to either more nuanced error functions or varying margins of error.

To the best of our knowledge, the ALICE Score is the first
competence estimator that is scalable to large models and datasets
and is generalizable to all classifiers, error functions, and
performance levels. Our method takes into account all three aspects of predictive uncertainty in order to accurately predict competence on all of
the models and datasets that it has encountered, regardless of the
stage of training. Further, it does not require any
out-of-distribution data to train on and can easily be interpreted
as a probability of model competence. It also provides
tunable parameters of \(\delta\), \(\epsilon\), and \(\mathcal{E}\)
allowing for a more flexible version of competence that can fit a
variety of users' needs.
\section{Evaluating Competence Estimators}
\label{sec:orgf69036b}
\subsection{Binary \(\delta-\epsilon\) Competence Classification}
\label{sec:org16f5dbb}
We consider the task of pointwise binary competence
classification. Given \(f(x)\) and \(\hat{f}(x)\), we can directly
calculate \(\mathcal{E}(f(x), \hat{f}(x))\) and thus the model's true
\(\delta\) competence on \(x\). Given a competence estimator, we can
then \emph{predict} if the model is \(\delta\) competent on \(x\), thus
creating a binary classification task parametrized by
\(\epsilon\). This allows us to use standard binary classification
metrics such as Average Precision (AP) across all recall values to
evaluate the competence estimator.

We note that the true model competence is nondecreasing as \(\delta\)
increases since we are strictly increasing the support. In
particular, we have that the model is truly incompetent with
respect to \(\mathcal{E}\) on all points when \(\delta = 0\), and the
model is truly competent with respect to \(\mathcal{E}\) on all
points as \(\delta \to \infty\) as long as \(\mathcal{E}\) is bounded
above. This makes it difficult to pick a single \(\delta\) that is
representative of the performance of the competence estimator on a
range of \(\delta\)'s. To mitigate this issue we report \emph{mean} AP
over a range of \(\delta\)'s, as this averages the
estimator's precision across these error tolerances.

Note that this metric only evaluates how well each estimator \emph{orders} the test points based on competence, and does not consider the actual value of the score. We test this since some competence estimators (e.g. TrustScore) only seek to \emph{rank} points based on competence and do not care what the magnitude of the final score is. As a technical detail, this means that we cannot parametrize the computation of Average Precision by $\epsilon$ (since some estimators don't output scores in the range [0, 1]), and must instead parametrize each estimator's AP computation separately by thresholding on that estimator's output.
\section{The ALICE Score: \(\delta-\epsilon\) competence estimation}
\label{sec:org68a3dad}
 We would like to determine whether or not the model is competent on
a point without knowledge of ground truth, as in a test-set scenario
where the user does not have access to the labels of a data
point. Formally, given a \(\delta\) and an input \(x\), we want to
estimate \(p(\mathcal{E}(f(x), \hat{f}(x)) < \delta | x)\).

We write \(p(\mathcal{E}(f(x), \hat{f}(x)) < \delta | x)\) as
\(p(\mathcal{E} < \delta | x)\), where \(\mathcal{E}\) is the random
variable that denotes the value of the \(\mathcal{E}\) function given a
point \(x\) and its label \(f(x)\). We begin by marginalizing over the
possible label values \(f(x) = c_j \in \mathcal{Y}\) (where $c_j$ is the one-hot label for class $j$):
\begin{align}
p(\mathcal{E} < \delta | x)
&= \sum_{c_j \in \mathcal{Y}} p(\mathcal{E} < \delta | c_j, x) p(c_j |
x)\\  &= \sum_{c_j \in \mathcal{\hat{Y}}\cap \mathcal{Y}} p(\mathcal{E} < \delta | c_j, x)
p(c_j | x)  + \sum_{c_j \in
\mathcal{\overline{\hat{Y}}} \cap \mathcal{Y}} p(\mathcal{E} < \delta | c_j, x) p(c_j | x) \label{new_label} \\
&=  \sum_{c_j \in
\mathcal{\hat{Y}}} p(\mathcal{E} < \delta | c_j, x) p(c_j | x)\label{alice_label}
\end{align} 
Note that the \(\mathcal{E}(c_j, \hat{f}(x))\) was defined to be
\(\infty\) when \(c_j \in \overline{\hat{\mathcal{Y}}} \cap \mathcal{Y}\)
(Definition \ref{error}), thus the rightmost summation in Equation \ref{new_label} is 0 for all
\(\delta\). Furthermore, since $\hat{\mathcal{Y}} \subseteq \mathcal{Y}$ (Definition \ref{error}) we have $\hat{\mathcal{Y}} \cap \mathcal{Y} = \hat{\mathcal{Y}}$ which gives the final equality.  
To explicitly capture \emph{distributional uncertainty}, we now
marginalize over the variable \(D\), which we define as the event that
\(x\) is in-distribution:

\begin{align}	
p(\mathcal{E} < \delta | x) &= \sum_{c_j \in \hat{\mathcal{Y}}} p(\mathcal{E} < \delta | c_j,
x) p(c_j | x) \nonumber \\
&= \sum_{c_j \in \hat{\mathcal{Y}}} p(\mathcal{E} < \delta | c_j, x, D)p(c_j| x,
D)p(D|x) + \sum_{c_j \in \hat{\mathcal{Y}}} p(\mathcal{E} < \delta | c_j, x, \overline{D})p(c_j |x, \overline{D})p(\overline{D}|x) \label{dist_margin}
\end{align}
Consider the rightmost summation in Equation \ref{dist_margin}. This represents
the probability that the model is competent on the point \(x\) assuming
that \(x\) is out-of-distribution. However, this term is intractable to
approximate due to \emph{distributional uncertainty}. Given only in-distribution training data, we assume that we
cannot know whether the model
will be competent on out-of-distribution test points. To mitigate this concern we lower bound the estimation by
setting this term to \(0\) --- this introduces the inductive bias
that the model is not competent on points that are
out-of-distribution. This simplification yields:
\begin{align}
p(\mathcal{E} < \delta | x) &\geq p(D | x) \sum_{c_j \in
\hat{\mathcal{Y}}} p(\mathcal{E} < \delta | c_j, x)
p(c_j | x, D) \label{alice_decomp}
\end{align}
This allows our estimate to err on the side of caution as we would
rather predict that the model is incompetent even if it is truly
competent compared to the opposite situation. We approximate each of
the terms in Equation \ref{alice_decomp} in turn.

\subsection{Approximating \(p(D|x)\)}
\label{sec:orgfd0cf1b}
This term computes the probability that a point \(x\) is
in-distribution. We follow a method derived from the state-of-the-art anomaly detector \citep{lee2018simple} to compute this term: For
each class \(j\) we fit a class-conditional Gaussian \(G_{j}\) to the set \(\{x \in
   \mathcal{X}_{train} :\hat{f}(x) = c_j\}\) where
\(\mathcal{X}_{train}\) is the training data. Given a test point \(x\)
we then compute the Mahalanobis distance \(d_j\) between \(x\)
and \(G_{j}\). In order to turn this distance into a probability, we
consider the empirical distribution \(\beta_{j}\) of possible
in-distribution distances by computing the distance of each
training point to the Gaussian \(G_{j}\), and then computing the survival function. We take the maximum value of the survival function across all $j$. This intuitively models the probability that the point
is in-distribution with respect to \emph{any} class. Explicitly, we have
\(p(D|x) = \max_j{1 - \text{CDF}_{\beta_{j}}(d_{j})}\). Note that
this term measures distribution shift, which closely aligns with
\emph{distributional uncertainty}.

\subsection{Approximating \(p(\mathcal{E} < \delta | x, c_j)\)}
\label{sec:org2c86f6a}
This term computes the probability that the error at the point
\(x\) is less than \(\delta\) given that the one-hot label is \(c_j\). We
directly compute \(\mathcal{E}(c_j, \hat{f}(x))\), then simply
check whether or not this error is less than \(\delta\).  Note that
this value is always \(1\) or \(0\) since it is the indicator
\(\mathbbm{1}[\mathcal{E}(c_j, \hat{f}(x)) < \delta]\), and that
this term estimates the difference between the predictions of \(f\)
and \(\hat{f}\), which aligns with \emph{model uncertainty}.

\subsection{Approximating \(p(c_j | x, D)\)}
\label{sec:orge705596}
This term computes the probability that a point \(x\) is of class
\(j\), given that it is in-distribution. To estimate this
class probability, we fit a transfer classifier at the given layer
and use its class-probability output, \(\hat{p}(c_j | x,
   D)\). Since the test points are assumed to be in-distribution, we can trust the output of the classifier as long as it is calibrated --- that is, for all \(x\) with
\(p(c_j | x) = p\), \(p\) of them belong to class \(j\). \citep{Niculescu-Mizil_platt} examines the calibration
of various classifiers, and shows that Logistic Regression (LR)
Classifiers are well calibrated. Random
Forests and Bagged Decision Trees are also calibrated
\citep{Niculescu-Mizil_platt}, however, we find that the choice of
calibrated classifier has little effect on the accuracy of our
competence estimator. Note that --- with a perfectly calibrated
classifier --- this term estimates the uncertainty inherent in the data
(e.g. a red/blue classifier will always be uncertain on purple
inputs due to class overlap), which closely
aligns with \emph{data uncertainty}.

\subsection{The ALICE Score}
\label{sec:org1dd92fa}

Putting all of these approximations together yields the ALICE
Score: 
\begin{align}
p(\mathcal{E}(f(x), \hat{f}(x)) < \delta | x) &\gtrapprox  \max_j{(1 - \text{CDF}_{\beta_{j}}}(d_{
  j})) \sum_{c_j \in \hat{\mathcal{Y}}} \mathbbm{1}[\mathcal{E}(\hat{f}(x), c_j) < \delta]\hat{p}(c_j|x, D) \label{alice_score}
  \end{align}

Note that the ALICE Score can be written at
layer \(l\) of a neural network by treating \(x\) as the activation of layer \(l\) in a
network and using those activations for the transfer classifiers
and the class conditional Gaussians.

We do not claim that the individual components of the ALICE Score
are optimal nor that our estimator is optimal --- we merely wish
to demonstrate that the ALICE framework of expressing competence
estimation according to Equation \ref{alice_score} is empirically
effective. 
\section{Experiments and Results}
\label{sec:org7a7ceed}
\subsection{Experimental Setup}
\label{sec:org5fd59d2}
We conduct a variety of experiments to empirically evaluate ALICE as
a competence estimator for classification tasks. We vary the model,
training times, dataset, and error function to show the robustness of
the ALICE Score to different variables. We compute metrics for competence prediction by simply using the
score as a ranking and thresholding by recall values to compare with
other scores that are neither \(\epsilon\)-aware nor calibrated, as
discussed in Section 4. The mean Average Precision is computed across
\(100\) \(\delta\)'s linearly spaced between the minimum and maximum of
the \(\mathcal{E}\) output (e.g. for cross-entropy we space \(\delta\)'s
between the minimum and the maximum cross-entropy error on a validation set). For all experiments, we compute ALICE scores
on the penultimate layer, as we empirically found this layer to
provide the best results --- we believe this is due to the penultimate layer having the most well-formed representations before the final predictions. We compare our method only with Trust Score and model confidence (usually the softmax score) since they apply to all models and do not require extraneous data. Further experimental details are provided
in Appendix A. 
\subsection{Predictive Uncertainty Experiments}
\label{sec:org6194613}
Since competence is a generalized form of confidence, and confidence
amalgamates all forms of predictive uncertainty, competence
estimators must account for these factors as well. We empirically
show that ALICE can accurately predict competence when encountering
all three types of predictive uncertainty --- note that we do not
claim that the ALICE framework perfectly disentangles these three
facets, merely that each term is essential to account for all forms
of predictive uncertainty. 

\begin{figure}
\begin{subfigure}{.5\textwidth}
\centering
\includegraphics[width=.8\linewidth]{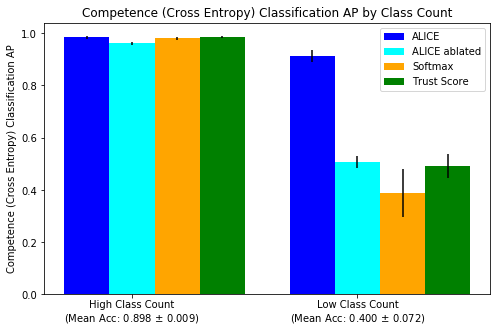}
\caption{mAP of competence scores ($\mathcal{E}$ = cross-entropy)}
\label{class_imb_a}
\end{subfigure}
\begin{subfigure}{.5\textwidth}
\centering
\includegraphics[width=.8\linewidth]{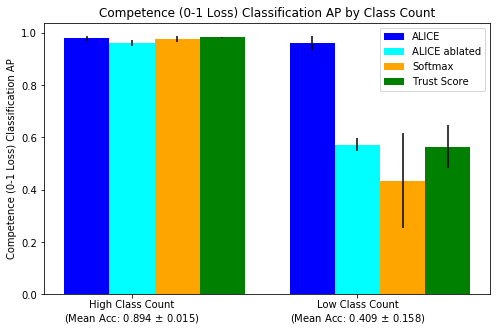}
\caption{mAP of competence scores ($\mathcal{E}$ = 0-1 error)}
\label{class_imb_b}
\end{subfigure}
\caption{Competence Scores on Class Imbalanced CIFAR10}
\label{class_imbalance}
\end{figure}

We first examine \emph{model uncertainty} by performing an ablation study
on both overfit and underfit classical models on DIGITS and VGG16
\citep{Simonyan14c} on CIFAR100 \citep{cifar100}. Details about these
models are in Appendix A. As expected, ALICE strongly outperforms the
other metrics in areas of over and underfitting and weakly
outperforms in regions where the network is trained well (Table \ref{tab:orgdd56792}). Further, we highlight a specific
form of \emph{model uncertainty} in Figure \ref{class_imbalance} by
performing the same ablation study on the common situation of
class-imbalanced datasets. We remove \(95\)\% of the training data for
the final \(5\) classes of CIFAR10 so that the model is poorly matched
to these low-count classes, thus introducing model
uncertainty. Figure \ref{class_imbalance} shows the mean Average
Precision (mAP) of competence prediction on the unmodified CIFAR10
test set after fully training VGG16 on the class-imbalanced CIFAR10
dataset. While all of the metrics perform similarly on the classes of
high count, neither softmax (orange) nor trust score (green) were
able to accurately predict competence on the low count classes. ALICE
(blue), on the other hand, correctly identifies competence on all
classes because ALICE considers \emph{model uncertainty}. We additionally show that omitting the term $p(\mathcal{E} < \delta | x, c_j)$ removes this capability, thus empirically showing that this term is necessary to perform accurate competence estimation under situations of model uncertainty.

\begin{table}[htbp]
\caption{\label{tab:orgdd56792}
mAP for Competence Prediction Under Model Uncertainty (\(\mathcal{E}\) = cross-entropy). VGG16 is tested on CIFAR100 while the other models are on DIGITS. (U) is underfit, (W) is well trained, and (O) is overfit. Ablated ALICE refers to ALICE without the $p(\mathcal{E} < \delta | x, c_j)$ terms. Hyperparameters for these trials are in Appendix A.}
\centering
\begin{tabular}{ccccccc}
\toprule
Model & Accuracy & Softmax & TrustScore & Ablated ALICE & ALICE\\
\midrule
MLP (U) & .121 \textpm{} .048 & .0486 \textpm{} .015 & .505 \textpm{} .27 & .0538 \textpm{} .031 & \textbf{.999 \textpm{} .0015}\\
MLP (W) & .898 \rpm .022 & .989 \rpm .005 & .929 \rpm .044 & .958 \rpm .042 & \textbf{.998 \rpm .001}\\
MLP (O) &  .097 \rpm .015 & .532 \rpm .062 & .768 \rpm .064  & .576 \rpm .033 & \textbf{.996 \rpm .003}\\
RF  (U) & .563 \textpm{} .078 & .824 \textpm{} .16 & .504 \textpm{} .33 & .290 \textpm{} .322 & \textbf{.999 \textpm{} .0011}\\
RF (W) &  .930 \rpm .019 & .998 \rpm .002 & .898 \rpm .025 & .923 \rpm .016 & \textbf{.999 \rpm .000}\\
SVM (U) & .630 \rpm .018 & .995 \rpm .003 & .626 \rpm .046  & .496 \rpm .069 & \textbf{1.00 \rpm .000}\\
SVM (W) &  .984 \rpm .009 & \textbf{1.00 \rpm .000} &   .931 \rpm .048 & .963 \rpm .038 & \textbf{1.00 \rpm .000}\\
SVM (O) & .258 \textpm{} .023 & .200 \textpm{} .16 & .215 \textpm{} .12 & .252 \textpm{} .16 & \textbf{.981 \textpm{} .028}\\
\midrule
VGG16 (U) & .0878 \textpm{} .0076 & .899 \textpm{} .014 & .292 \textpm{} .049 & .0369 \textpm{} .0041 & \textbf{.913 \textpm{} .012}\\
VGG16 (W) & .498 \textpm{} .012 & .975 \textpm{} .013 & .604 \textpm{} .104 & .0863 \textpm{} .0071 & \textbf{.978 \textpm{} .0082}\\
VGG16 (O) & .282 \textpm{} .15 & .659 \textpm{} .024 & .665 \textpm{} .0080 & .257 \textpm{} .018 & \textbf{.738 \textpm{} .019}\\
\bottomrule
\end{tabular}
\end{table}
While Figure \ref{class_imbalance} and Table \ref{tab:orgdd56792} show ALICE's performance under situations of high \emph{model uncertainty}, we show ALICE's performance under situations of \emph{distributional uncertainty} in Table
\ref{tab:org046e03f}. First we define a \emph{distributional competence} error
function: $$\mathcal{E_D}(f(x), \hat{f}(x)) = \begin{cases} 0 & f(x)
\in \hat{\mathcal{Y}} \\ 1 & f(x) \notin \hat{\mathcal{Y}}
\end{cases}$$This function is simply an indicator as to whether or not
the true label of a point is in the predicted label space. We fully train ResNet32 on the
unmodified CIFAR10 training set. We then compute competence scores with respect to $\mathcal{E}_D$ on a
test set with varying proportions of SVHN \citep{svhn} (out-of-distribution)
and CIFAR10 (in-distribution) data. In
this case \(\mathcal{Y} = \mathcal{Y}_{\text{CIFAR}} \cup
\mathcal{Y}_{\text{SVHN}}\) but \(\mathcal{\hat{Y}} =
\mathcal{Y}_{\text{CIFAR}}\), thus \(\mathcal{E}_D\) is \(1\) on SVHN
points and \(0\) on CIFAR points. Table \ref{tab:org046e03f} shows that both
softmax and ALICE without the \(p(D|x)\) term perform poorly on
distributional competence. In contrast, both the full ALICE score
and Trust Score are able to estimate distributional competence in all
levels of distributional uncertainty --- this is expected since ALICE
contains methods derived from a state-of-the-art anomaly detector
\citep{lee2018simple} and Trust Score considers distance to the training data. Note that this construction of the distributional competence function is a clear example of how the general notion of competence can vary tremendously depending on the task at hand, and ALICE is capable of predicting accurate competence estimation for any of these notions of competence.

\begin{table}[htbp]
\caption{\label{tab:org046e03f}
mAP for Competence Prediction Under Distributional Uncertainty (\(\mathcal{E} = \mathcal{E_D}\)).}
\centering
\begin{tabular}{ccccccc}
\toprule
CIFAR/SVHN Proportion & Softmax & TrustScore & Ablated ALICE & ALICE\\
\midrule
10/90 & .458 \rpm 0.056 & .518 \rpm 0.039  & .100 \rpm 0.000 & \textbf{.868 \rpm 0.014}\\
30/70 &  .693 \rpm 0.034 & .721 \rpm 0.026  & .300 \rpm 0.000 & \textbf{.946 \rpm 0.007}\\
50/50 &  .816 \rpm 0.020 & .833 \rpm 0.015  & .500 \rpm 0.000 & \textbf{.970 \rpm 0.003}\\
70/30 &  .901 \rpm 0.010 & .910 \rpm 0.008  & .700 \rpm 0.000 & \textbf{.985 \rpm 0.002}\\
90/10 &  .970 \rpm 0.003 & .972 \rpm 0.002  & .900 \rpm 0.000 & \textbf{.997 \rpm 0.001}\\
\bottomrule
\end{tabular}
\end{table}
\begin{figure}
\centering
\includegraphics[width=.8\linewidth, height=.25\linewidth]{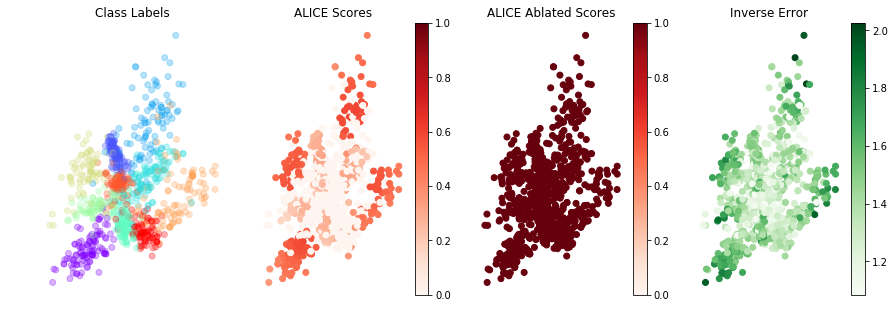}
\caption{Competence Visualization on CIFAR10 ($\delta = .001,
\mathcal{E} =$ cross-entropy). Points are projected to
two dimensions with Neighborhood Component Analysis. From left to
right, figures are colored by the class label, ALICE Score,
Ablated ALICE Score, and inverse error (so darker colors imply competence).}
\label{comp_vis}
\end{figure}
We examine ALICE's capturing of \emph{data uncertainty} by observing
competence predictions in areas of class overlap in Figure
\ref{comp_vis}. Here we trained VGG16 on CIFAR10 \citep{cifar10} and
visualized competence scores with respect to cross-entropy. Note that
the competence scores are very low in areas of class overlap, and that
these regions also match with areas of high error. Additional experiments with varying models, error functions, and levels of uncertainty are provided in Appendix B. 
\subsection{Calibration Experiments}
\label{sec:org50bc77b}
\begin{figure}[ht]
\centering

\begin{subfigure}{.18\textwidth}
\includegraphics[height=3.35cm]{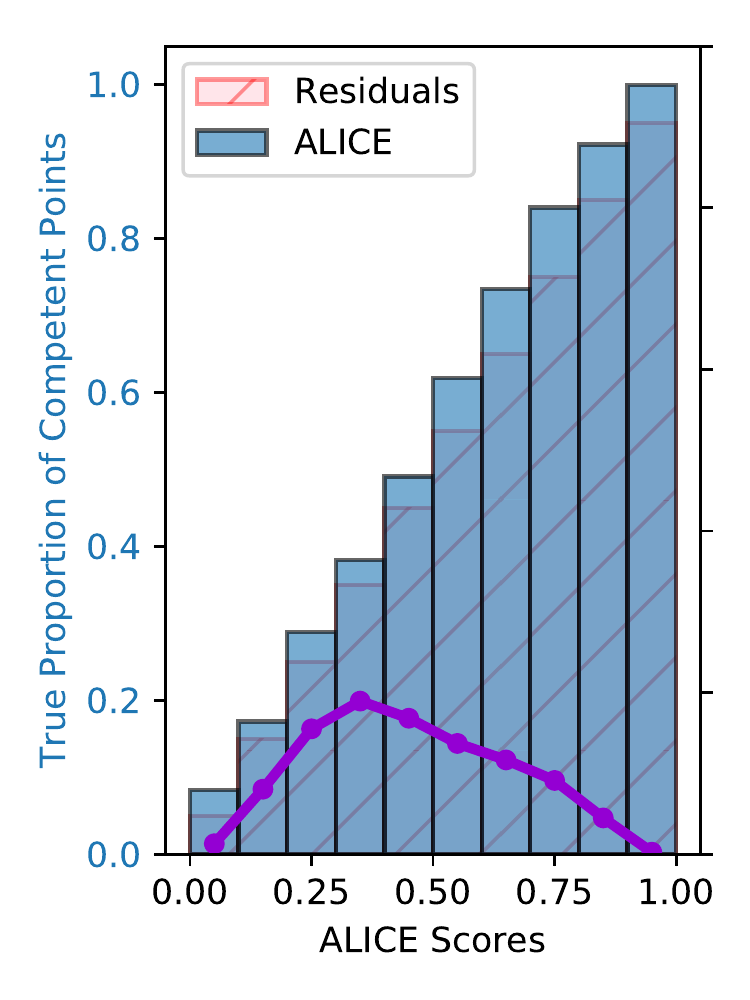}
\vspace{-2em}
\caption{$\mathcal{E}_{0\text{-}1}$ ($1$)}
\end{subfigure}
\hspace{-1.35em}
\begin{subfigure}{.14\textwidth}
\includegraphics[height=3.35cm]{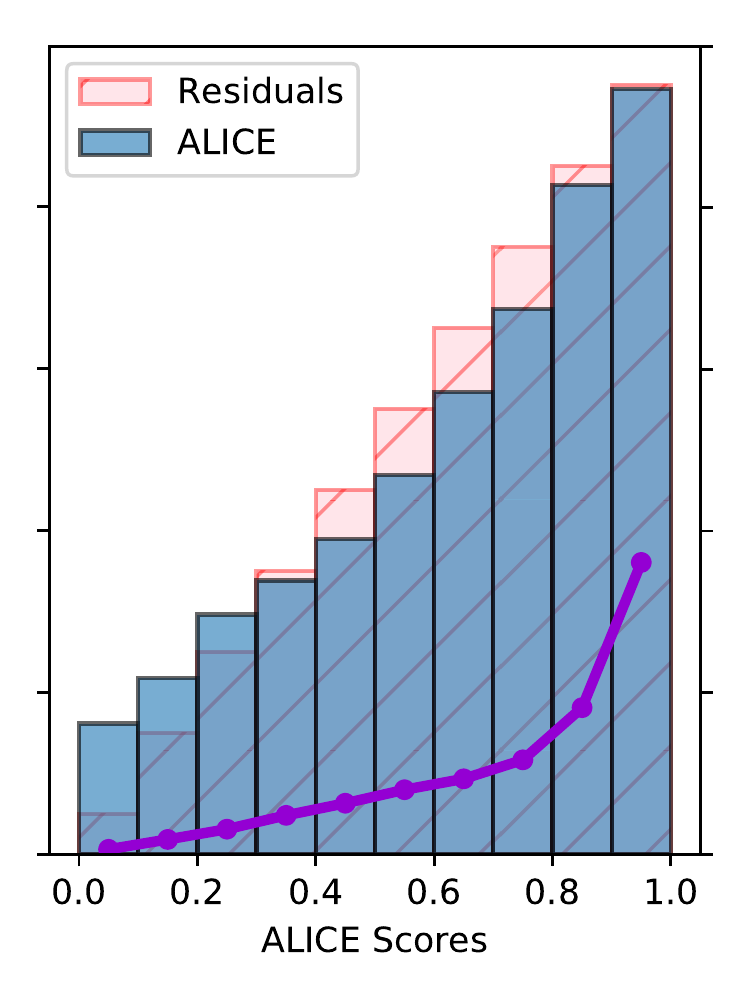}
\vspace{-2em}
\caption{$\mathcal{E}_{0\text{-}1}$ ($5$)}
\end{subfigure}
\hspace{.5em}
\begin{subfigure}{.14\textwidth}
\includegraphics[ height=3.35cm]{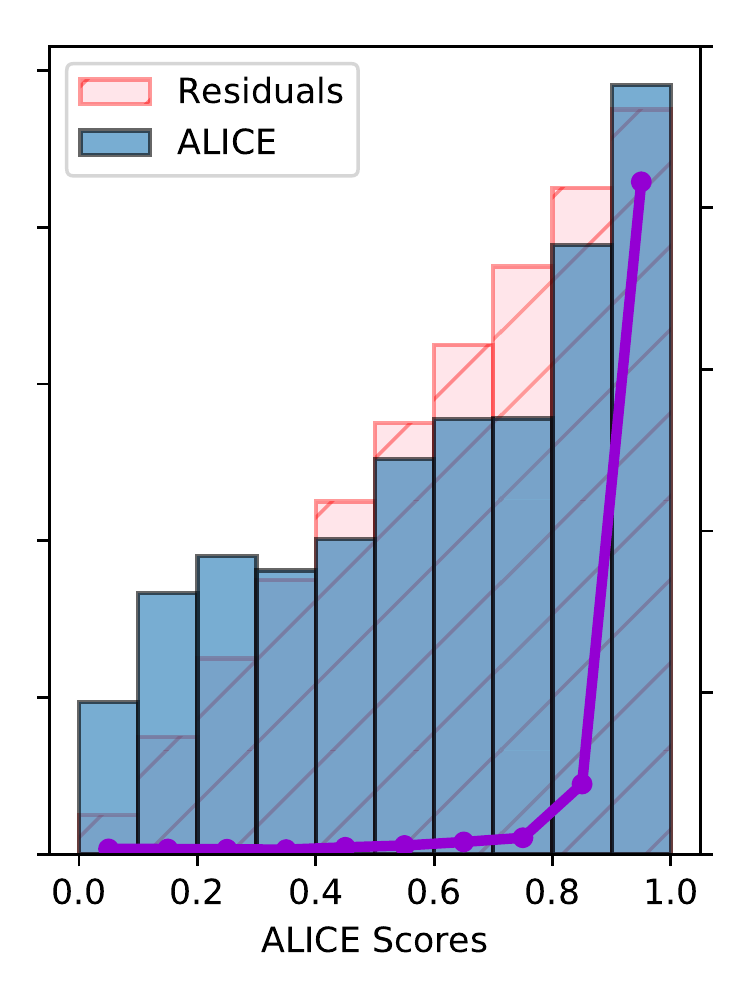}
\vspace{-2em}
\caption{$\mathcal{E}_{0\text{-}1}$ ($50$)}
\end{subfigure}
\hspace{.5em}
\begin{subfigure}{.14\textwidth}
\includegraphics[height=3.35cm]{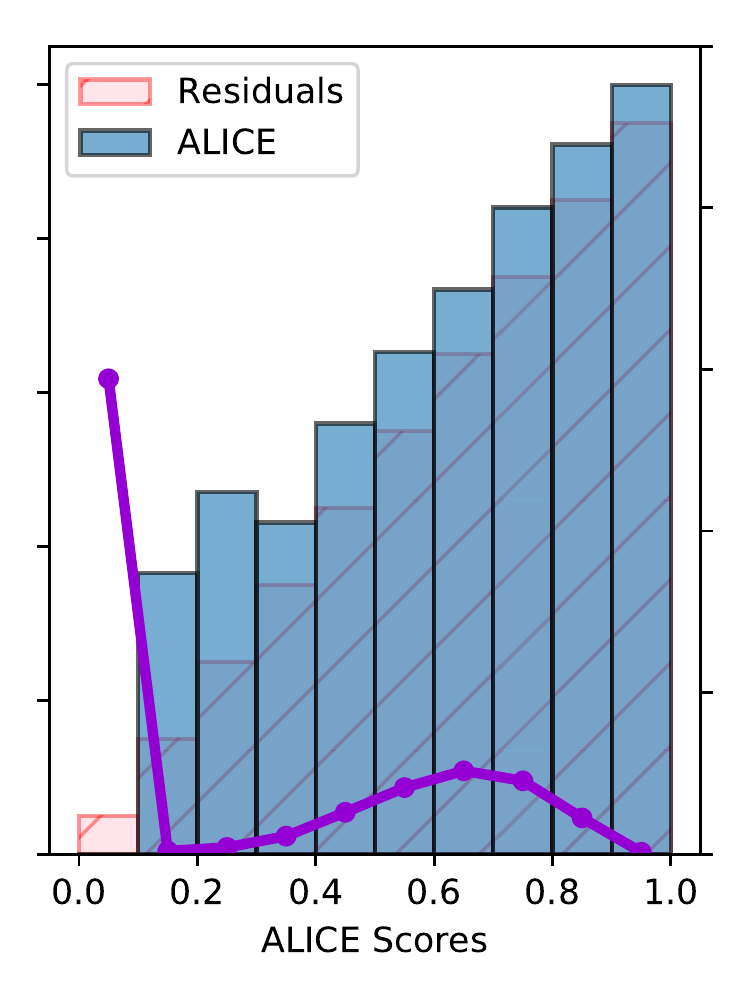}
\vspace{-2em}
\caption{$\mathcal{E}_{\text{xent}}$ ($1$)}
\end{subfigure}
\hspace{.5em}
\begin{subfigure}{.14\textwidth}
\includegraphics[ height=3.35cm]{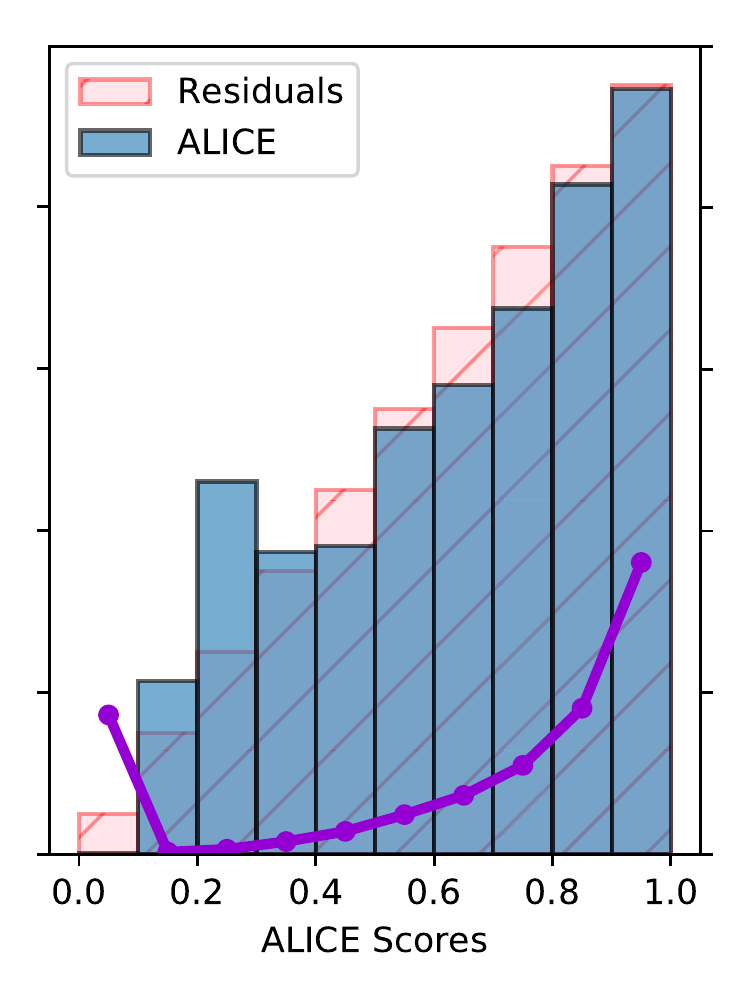}
\vspace{-2em}
\caption{$\mathcal{E}_{\text{xent}}$ ($5$)}
\end{subfigure}
\hspace{.5em}
\begin{subfigure}{.18\textwidth}
\includegraphics[height=3.35cm]{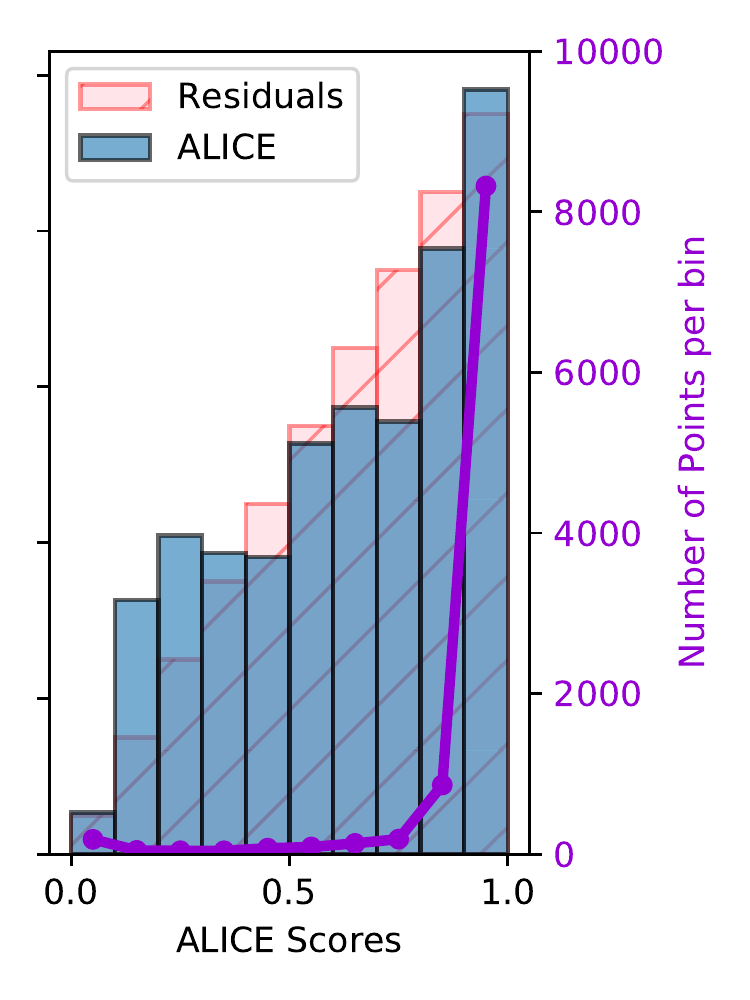}
\vspace{-2em}
\caption{$\mathcal{E}_{\text{xent}}$ ($50$)}
\end{subfigure}
\caption{\label{calibration}ALICE score calibration of ResNet32 trained on CIFAR10, with various error functions and stages of training. The captions show the error functions and number of epochs trained.}
\end{figure}

While the previous experiments show the ability of ALICE to \emph{rank} points according to competence, we now show the
interpretability of the ALICE score through calibration curves. Note that we are not attempting to interpet or explain \emph{why} the model has made the decision that it has, we simply aim to show that the ALICE score matches its semantic meaning: for all points with ALICE score of $p$, we expect $p$ of them to be truly competent. To show this, we train ResNet32 on CIFAR100 and compute ALICE scores at various stages of training and for different error functions (we use $\delta$ = 0.2 when computing competence for $\mathcal{E}_{\text{xent}}$. We bin the ALICE scores into tenths ([0.0 - 0.1), [0.1 - 0.2), ..., [0.9, 1.0)) and plot the true proportion of competent points for each bin as a histogram. Note that a perfect competence estimation with infinite data would result in these histograms roughly resembling a $y = x$ curve. We visualize the difference between our competence estimator and perfect competence estimation by showing these residuals as well as the number of points in each bin in Figure \ref{calibration}. Note that ALICE is relatively well-calibrated at all stages of training and for all error functions tested --- this result shows that one can \emph{interpret} the ALICE score as an automatically calibrated probability that the model is competent on a particular point. This shows that not
only does the ALICE Score \emph{rank} points accurately according to
their competence but it also rightfully assigns the
correct probability values for various error functions and at all stages of training.

\section{Conclusions and Future Work}
\label{sec:orge067940}
In this work we present a new, flexible definition of competence. Our
definition naturally generalizes the notion of confidence by allowing a variety of error functions as well as risk and correctness thresholds in order to construct a definition that is
tunable to an end-user's needs. We also develop the ALICE Score, an
accurate layerwise interpretable competence estimator for
classifiers. The ALICE Score is not only applicable to any classifier
but also outperforms the state-of-the-art in competence
prediction. Further, we show that the ALICE Score is robust to
out-of-distribution data, class imbalance and poorly trained models
due to our considerations of all three facets of predictive
uncertainty.

The implications of an accurate competence estimator are far
reaching. For instance, future work could include using the ALICE
Score to inform an Active Learning acquisition function by labeling
points that a model is least competent on. One could also examine a
network more closely by performing feature visualization or finding
prototypes in areas of low competence, as this would elucidate which
features are correlated with incompetence. This is particularly useful
since the ALICE Score can be computed layerwise in order to find both
low and high level features that the model is not competent
on. Competence estimators could also be used as test and evaluation
metrics when a model is deployed to detect both distributional shift
and classification failure.

Future work will focus on extending the ALICE Score to supervised tasks other
than classification such as object detection, segmentation, and regression. Additionally, because many of the components of the ALICE
Score are state-of-the-art for detecting adversarial examples, we
expect that the ALICE Score would also be able to detect adversarial
samples and assign them low competence, though we have not tested this
explicitly. Further research will also include better approximations
of the terms in the ALICE Score to improve competence
estimation. Finally, we plan to explore different methods to ensemble
the layerwise ALICE Scores into an overall ALICE Score for the model
and determine whether or not that improves performance compared to the
layerwise ALICE Scores.

\newpage

\subsubsection*{Acknowledgements}
The authors would like to thank the JHU/APL Internal Research and Development (IRAD) program for funding this 
research.
 
\bibliography{alice_reborn}
\newpage
\section{Appendix A: Experimental Details}
\label{sec:org5b4e5e4}
For the CIFAR experiments we initialize VGG16 with pretrained
imagenet weights and for the DIGITS experiments we use classical
machine learning models such as Support Vector Machines (SVMs), Random
Forests (RFs), Multilayer Perceptrons (MLPs), and Logistic Regression (LR). VGG16 and
LeNET are trained with learning rates of \(1e-4\) and \(3e-4\)
respectively, and default regularization parameters. The deep models
are trained on a single NVIDIA 1080TI GPU, while the classical models
are all trained on the CPU. For all experiments we perform ten trials
and report the mean and standard deviation. The default scikit-learn
parameters are used for SVM's, LR, RF, MLP unless stated otherwise
with the exception of the SVM regularization parameter \(C\) which we
set to \(0.01\). We use an MLP with one hidden layer of \(10\) neurons. The grid search for ALICE's transfer classifier regularization parameter is over a logspace from $-5$ to $5$.

For the model uncertainty experiments we have trained the MLP for 1
iteration to ensure underfitting. We set the RF to have max depth = 1
and 20 estimators to also ensure the model will poorly match the
data. The SVM has a degree \(5\) polynomial kernel and no
regularization to ensure overfitting to the data. The well-trained
VGG16 network has a regularization parameter of 1e+8 on the final two
layers and is fully trained. The underfit and overfit networks both
have a regularization parameter of \(0\), but are trained for \(1\) and
\(100\) epochs respectively.

Each dataset is randomly split (unless a split already exists) into
an \(80-10-10\) split of train-test-validation, and each model is
trained until the performance on a validation set is maximized unless
explicitly stated otherwise. The deep models are all trained with
respect to cross-entropy loss, while the classical models are fitted
with their respective loss functions.
\section{Appendix B: Further Competence Prediction Experiments}
\label{sec:orgffe99b7}
Here we show competence prediction on various models with respect to
different error functions on the DIGITS and CIFAR100 datasets. Note
that ALICE consistently performs well regardless of model type or
model accuracy, and significantly outperforms both Trust Score and
model confidence on almost every model except for SVM with a
polynomial kernel, where it loses by a statistically insignificant
amount. Note that as the model accuracy increases, the improvement of
ALICE over the other methods decreases --- this is because model
uncertainty is decreasing, which alleviates the failures of methods
that do not consider model uncertainty.

\subsection{Competence Prediction (Cross-Entropy)}
\label{sec:org839313a}
\begin{center}
\captionof{table}{\label{tab:org4cd92f1}
mAP for Competence Prediction on DIGITS (\(\mathcal{E}\) = cross-entropy)}
\begin{tabular}{ccccc}
\toprule
Model & Accuracy & Softmax & TrustScore & ALICE\\
\midrule
SVM (RBF) & .147 \textpm{} .032 & .414 \textpm{} .15 & .346 \textpm{}.086 & \textbf{.989 \textpm{} .0078}\\
SVM (Poly) & .988 \textpm{} .007 & \textbf{1.00 \textpm{} .00027} & .949 \textpm{} .0073 & .999 \textpm{} 0.0011\\
SVM (Linear) & .971 \textpm{} .011 & \textbf{.999 \textpm{} .00066} & .951 \textpm{} .0094 & \textbf{.999 \textpm{} .00084}\\
RF & .928 \textpm{} .013 & .998 \textpm{} .0014 & .876 \textpm{} .013 & \textbf{1.00 \textpm{} .00048}\\
MLP (5 Iterations) & .158 \textpm{} .056 & .217 \textpm{} .13 & .579 \textpm{} .12 & \textbf{.966 \textpm{} .047}\\
MLP (200 Iterations) & .925 \textpm{} .017 & .988 \textpm{} .0087 & .963 \textpm{} .014 & \textbf{.998 \textpm{} .0017}\\
LR & .946 \textpm{} .017 & .995 \textpm{} .0029 & .977 \textpm{} .0041 & \textbf{.998 \textpm{} .0015}\\
\bottomrule
\end{tabular}
\end{center}
\begin{center}
\captionof{table}{\label{tab:orgdde6482}
mAP for Competence Prediction on CIFAR100 (\(\mathcal{E}\) = cross-entropy)}
\begin{tabular}{lrrrrl}
\toprule
Model & Accuracy & Softmax & Trust Score & ALICE\\
\midrule
ResNet50 (1 epoch) & .074 & .537 & .346 & \textbf{.723}\\
ResNet50 (5 epochs) & .293 & .895 & .715 & \textbf{.925}\\
ResNet50 (30 epochs) & .421 & .766 & .776 & \textbf{.828}\\
ResNet50 (100 epochs) & .450 & .795 & .807 & \textbf{.809}\\
\bottomrule
\end{tabular}
\end{center}

\subsection{Competence Prediction (MSE)}
\label{sec:orgae4d2ba}
\begin{center}
\captionof{table}{\label{tab:org166dc7a}
mAP for Competence Prediction on DIGITS (\(\mathcal{E}\) = mean-squared-error)}
\begin{tabular}{lllll}
\toprule
Model & Accuracy & Softmax & TrustScore & ALICE\\
\midrule
SVM (RBF) & .147 \textpm{} .032 & .394 \textpm{} .066 & .361 \textpm{} .046 & \textbf{.985 \textpm{} .011}\\
SVM (Poly) & .988 \textpm{} .007 & .999 \textpm{} .0018 & .990 \textpm{} .0045 & \textbf{.998 \textpm{} 0.0012}\\
SVM (Linear) & .971 \textpm{} .011 & 1.00 \textpm{} .00065 & .994 \textpm{} .0037 & \textbf{.999 \textpm{} .0013}\\
RF & .928 \textpm{} .013 & .996 \textpm{} .0016 & .956 \textpm{} .012 & \textbf{.999 \textpm{} .00034}\\
MLP (5 Iterations) & .158 \textpm{} .056 & .384 \textpm{} .11 & .746 \textpm{} .049 & \textbf{.992 \textpm{} .015}\\
MLP (200 Iterations) & .925 \textpm{} .017 & .986 \textpm{} .0069 & .985 \textpm{} .012 & \textbf{.997 \textpm{} .0027}\\
LR & .946 \textpm{} .017 & .995 \textpm{} .0025 & .989 \textpm{} .0051 & \textbf{.998 \textpm{} .0015}\\
\bottomrule
\end{tabular}
\end{center}

\subsection{Competence Prediction (\(\mathcal{E}\) = 0-1 error)}
\label{sec:org464fee2}
\begin{table}[htbp]
\caption{\label{tab:orgcb323fa}
AP for Competence Prediction on DIGITS (\(\mathcal{E}\) = 0-1 error)}
\centering
\begin{tabular}{lllll}
\toprule
Model & Accuracy & Softmax & TrustScore & ALICE\\
\midrule
SVM (RBF) & .147 \textpm{} .032 & .142 \textpm{} .27 & .106 \textpm{} .038 & \textbf{.983 \textpm{} .020}\\
SVM (Poly) & .988 \textpm{} .007 & \textbf{.999 \textpm{} .0013} & \textbf{.999 \textpm{} .00057} & \textbf{.999 \textpm{} .00042}\\
RF & .928 \textpm{} .013 & .994 \textpm{} .0034 & \textbf{1.00 \textpm{} .00050} & .998 \textpm{} .0015\\
SVM (Linear) & .971 \textpm{} .011 & \textbf{.999 \textpm{} .0011} & \textbf{.999 \textpm{} .00091} & .997 \textpm{} .0012\\
MLP (5 Iterations) & .158 \textpm{} .056 & .178 \textpm{} .094 & \textbf{.996 \textpm{} .0036} & .984 \textpm{} .014\\
MLP (200 Iterations) & .925 \textpm{} .017 & .981 \textpm{} .014 & \textbf{.999 \textpm{} .00037} & \textbf{.999 \textpm{} .0014}\\
LR & .946 \textpm{} .017 & .994 \textpm{} .0027 & \textbf{.999 \textpm{} .00038} & .996 \textpm{} .0017\\
\bottomrule
\end{tabular}
\end{table}
\begin{table}[htbp]
\caption{\label{tab:orgc9e623b}
AP for Competence Prediction on CIFAR100 (\(\mathcal{E}\) = 0-1 error)}
\centering
\begin{tabular}{lrrrl}
\toprule
Model & Accuracy & Softmax & TrustScore & ALICE\\
\midrule
VGG16 (1 epoch) & .209 & .514 & .654 & \textbf{.696}\\
VGG16 (5 epochs) & .323 & .670 & \textbf{.762} & .756\\
VGG16 (30 epochs) & .513 & .845 & .867 & \textbf{.873}\\
VGG16 (100 epochs) & .536 & .803 & .863 & \textbf{.871}\\
\bottomrule
\end{tabular}
\end{table}

\subsection{Histogram of ALICE Scores on in and out-of-distribution data}
\label{sec:org05f66ae}
\begin{figure}[ht]
\begin{subfigure}{.5\textwidth}
\centering
\includegraphics[width=.8\linewidth]{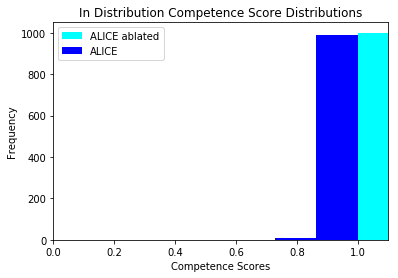}
\caption{Competence scores on MNIST test set}
\label{ood_cifar}
\end{subfigure}
\begin{subfigure}{.5\textwidth}
\centering
\includegraphics[width=.8\linewidth]{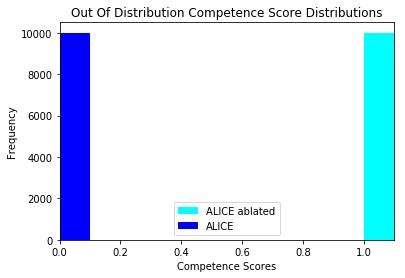}
\caption{Competence scores on CIFAR100 test set}
\label{ood_mnist}
\end{subfigure}
\caption{Competence Scores ($\mathcal{E} = \mathcal{E_D}$) on In-Distribution (MNIST) and
Out-of-Distribution (CIFAR100) Data for LeNet trained on MNIST. ALICE Ablated refers to ALICE without the
$p(D|x)$ term. Note how both the ALICE Scores and the ablated ALICE scores are both very high on the in-distribution examples; however, only the unablated ALICE scores are rightfully low when the model sees images from CIFAR100.} 
\label{ood_data}
\end{figure}

\begin{table}[htbp]
\caption{\label{class_overlap}
mAP for Competence Prediction Under Class Overlap (\(\mathcal{E} = \mathcal{E}_{0-1}\)). The datasets are synthetic datasets designed to show class overlap. Given a parameter $z$, we construct the dataset $D_z$ as follows. Class $0$ is a uniform distribution $U(-5, z)$, and class $1$ is a uniform distribution $U(-z, 5)$. Shifting $z$ from $0$ to $5$ yields different class overlap percentages. For the training set, we randomly generate $1000$ points from each class distribution; for the test and validation sets we randomly generate $100$ points from each class. We train a Logistic Regression Model and compute competence prediction scores with $\mathcal{E} = \mathcal{E}_{0-1}$. Ablated ALICE refers to ALICE without the $p(c_j | x, D)$ term. In contrast to the full ALICE score, ALICE ablated is unable to accurately predict competence in situations of class overlap. Additionally note that when there is 100 percent class overlap the model is randomly guessing, thus pointwise competence is also random.}
\centering
\begin{tabular}{cccccc}
\toprule
Overlap Percentage & Accuracy & Softmax & TrustScore & Ablated ALICE & ALICE\\
\midrule
0.00 & 1.00 \textpm{} 0.0 & \textbf{1.00 \textpm{} 0.0} & \textbf{1.00\textpm{} 0.0} & \textbf{1.00 \textpm{} 0.0} & \textbf{1.00 \textpm{} 0.0}\\
0.10 & .945 \textpm{} 0.0 & \textbf{.998 \textpm{} .000021} & \textbf{.998\textpm{} 0.0} & .945 \textpm{} 0.0 & \textbf{.998 \textpm{} 0.0}\\
0.25 & .865 \textpm{} 0.0 & .986 \textpm{} .000015 & 982\textpm{} 0.0 & .865 \textpm{} 0.0 & \textbf{.987 \textpm{} 0.0}\\
0.50 & .730 \textpm{} 0.0 & \textbf{.960 \textpm{} .000020} & .948 \textpm{} 0.0 & .730 \textpm{} 0.0 & \textbf{.960 \textpm{} 0.0}\\
0.75 & .625 \textpm{} 0.0 & \textbf{.862 \textpm{} 0.0} & .823\textpm{} 0.0 & .625 \textpm{} 0.0 & \textbf{.862 \textpm{} 0.0}\\
1.00 & .535 \textpm{} 0.0 & .499 \textpm{} 0.0 & .530\textpm{} 0.0 & \textbf{.535 \textpm{} 0.0} & .500 \textpm{} 0.0\\
\bottomrule
\end{tabular}
\end{table}

\begin{figure}[htbp]
\centering
\includegraphics[width=.9\linewidth]{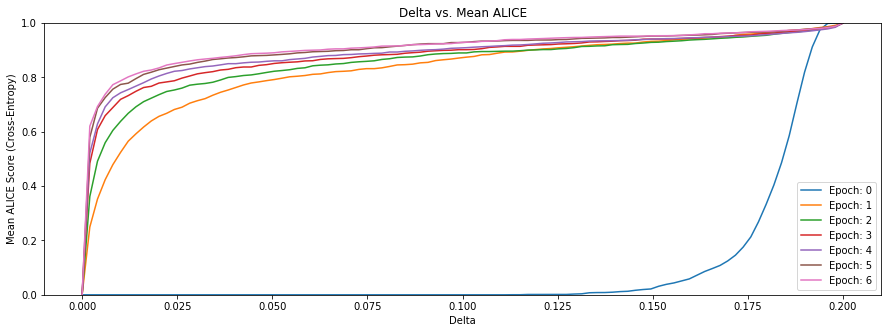}
\caption{\label{fig:orgb1cf2f2}
\(\delta\) vs mean ALICE Score across all points on MNIST ($\mathcal{E}$ = cross-entropy).}
\end{figure}
\begin{figure}[htbp]
\centering

\includegraphics[width=.9\linewidth]{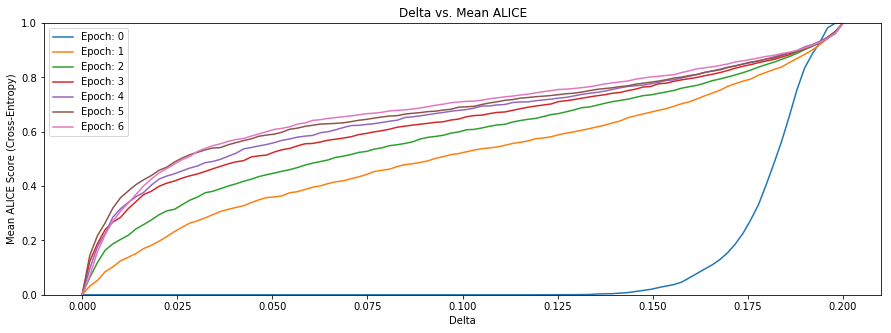}
\caption{\label{cifar_calibration}
\(\delta\) vs mean ALICE Score across all points on CIFAR10 ($\mathcal{E}$ = cross-entropy).}
\end{figure}

\clearpage
\vspace*{10mm}
\section{Appendix C: Competence Visualization}
\label{sec:orgce78c61}
\begin{figure}[h]
\begin{subfigure}{\textwidth}
\centering
\includegraphics[width=.8\linewidth, height=.25\linewidth]{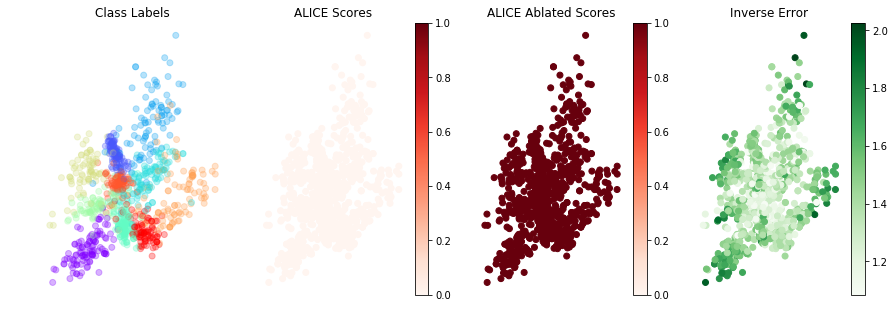}
\caption{$\delta = 0$}
\label{delta_0}
\end{subfigure}
\begin{subfigure}{\textwidth}
\centering
\includegraphics[width=.8\linewidth, height=.25\linewidth]{./pictures/1.png}
\caption{$\delta = 1e-3$}
\label{delta_14}
\end{subfigure}
\begin{subfigure}{\textwidth}
\centering
\includegraphics[width=.8\linewidth, height=.25\linewidth]{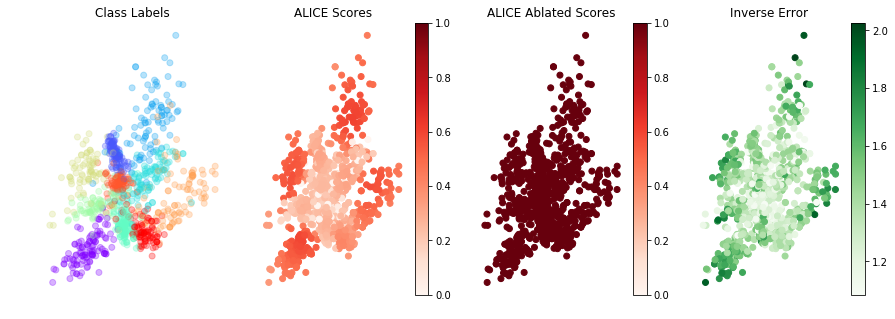}
\caption{$\delta = 1e-1$}
\label{delta_12}
\end{subfigure}
\begin{subfigure}{\textwidth}
\centering
\includegraphics[width=.8\linewidth, height=.25\linewidth]{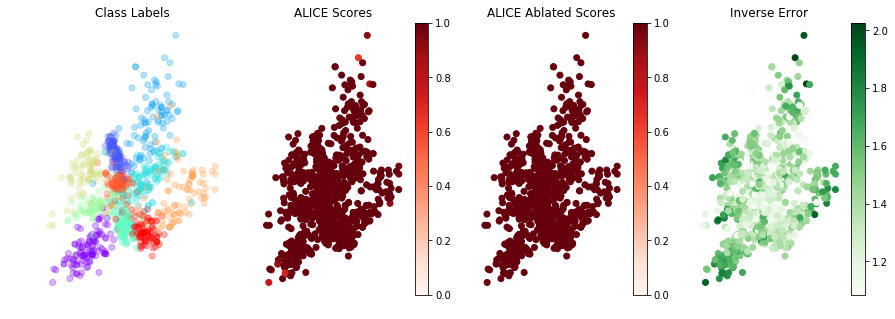}
\caption{$\delta = 2e-1$}
\label{delta_2}
\end{subfigure}
\caption{Competence Scores at varying \(\delta\)'s on the CIFAR10
Dataset with $\mathcal{E} = $cross-entropy. Points were projected to two dimensions with Neighborhood
Component Analysis. The left images are colored based on the actual
class label. The middle images are colored based on the predicted
\(\delta\)-competence at that specific $\delta$. ALICE Ablated refers to ALICE with $p(c_j | x_i, D)$ removed. Finally, the right
images show the inverse pointwise true error. Note how points increase in their ALICE Score as the error threshold increases. When $\delta = 0.2$ (the max value of $\mathcal{E}$ in the val set), the model is rightfully considered competent on nearly all points, and when $\delta = 0$ (the absolute min value of $\mathcal{E}$) the model is considered incompetent on nearly all points, as desired. }
\label{data_uncertainty}
\end{figure}
\end{document}